\documentclass[runningheads]{llncs}
\usepackage[T1]{fontenc}
\usepackage{amsmath,amsfonts}
\usepackage{algorithmic}
\usepackage{algorithm}
\usepackage{wrapfig}
\usepackage{graphicx}
\usepackage{caption}
\usepackage{subcaption}

\usepackage[utf8]{inputenc} 
\usepackage[T1]{fontenc}    
\usepackage{hyperref}       
\usepackage{url}            
\usepackage{booktabs}       
\usepackage{amsfonts}       
\usepackage{nicefrac}       
\usepackage{microtype}      
\usepackage{xcolor}         
\DeclareMathOperator*{\argmax}{arg\,max}
\DeclareMathOperator*{\argmin}{arg\,min}

\usepackage{todonotes}
\usepackage{graphicx}
\begin{document}
\title{Regularization of the policy updates for stabilizing Mean Field Games}
%
%

\author{Talal Algumaei\inst{1},
Ruben Solozabal\inst{1},
Reda Alami\inst{2},\\ Hakim Hacid\inst{2}, Merouane Debbah\inst{2} and Martin Tak\'a\v{c}\inst{1} }

\authorrunning{T. Algumaei et al.}

\institute{Mohamed bin Zayed University of Artificial Intelligence, Masdar City, UAE
\email{\{name.surname\}@mbzuai.ac.ae}\\
\and
Technology Innovation Institute, Masdar City, UAE\\
\email{\{name.surname\}@tii.ae}}

\maketitle              
\begin{abstract}

This work studies non-cooperative Multi-Agent Reinforcement Learning (MARL) where multiple agents interact in the same environment and whose goal is to maximize the individual returns. Challenges arise when scaling up the number of agents due to the resultant non-stationarity that the many agents introduce. In order to address this issue, Mean Field Games (MFG) rely on the symmetry and homogeneity assumptions to approximate games with very large populations. Recently, deep Reinforcement Learning has been used to scale MFG to games with larger number of states. Current methods rely on smoothing techniques such as averaging the q-values or the updates on the mean-field distribution. This work presents a different approach to stabilize the learning based on proximal updates on the mean-field policy. We name our algorithm \textit{Mean Field Proximal Policy Optimization (MF-PPO)}, and we empirically show the effectiveness of our method in the OpenSpiel framework.\footnote{This preprint has not undergone peer review or any post-submission improvements or corrections. The Version of Record of this contribution is published in PAKDD2023, and will be available online.}

\end{abstract}


\section{Introduction}



Despite the recent success of Reinforcement Learning (RL) in learning strategies in games (e.g., the game of Go~\cite{silver2016mastering}, Chess~\cite{schrittwieser2020mastering} or Starcraft~\cite{vinyals2019grandmaster}), learning in games with a large number of players is still challenging. Independent Learning leads to instabilities due to the fact that the environment becomes non-stationary. Alternatively, learning centralised policies can be applied to handle coordination problems and avoid the non-stationarity. However, centralised learning is hard to scale, as the joint action space grows exponentially with the number of agents. Many works in Multi-Agent Reinforcement Learning (MARL) have succeeded in decomposing the objective function into individual contributions~\cite{son2019qtran}, although this is also intractable when the number of agents is large. In this sense, mean field theory addresses large population games by approximating the distribution of the players. An infinite population of agents is represented by a continuous distribution of identical players that share the same behaviour. This reduces the learning problem to a representative player interacting with the representation of the whole population. 

This work in particular focuses on learning in Mean Field Games (MFG), non-cooperative games in which many agents act independently to maximise their individual reward, and the goal is to reach the Mean Field Nash Equilibrium (MFNE). Learning in MFG is not an easy task as most of the problems do not have an analytical solution. Traditionally numerical methods have been used to address these problems~\cite{lauriere2021numerical}; nonetheless, these methods do not scale well. In this sense, numerous game theory approaches have been brought into MFG. A classical algorithm is the Banach-Picard (BP)~\cite{huang2006large} algorithm, which uses a fixed-point iteration method to interactively update the population's behaviour based on the best response of a single representative agent against the mean-field distribution. However, acting in a best response to other agents might cause the others to actuate in the same way, leading to instabilities in the learning (referred to as the \textit{curse of many agents} in game-theory~\cite{sonu2017decision}). In practice, smoothing techniques derived from optimization theory are used to guarantee the convergence of these algorithms under reasonable assumptions~\cite{perolat2021scaling}.

More recently, deep RL has been introduced to scale MFG to games with larger state spaces~\cite{perolat2021scaling}. Nevertheless, traditional approaches cannot be directly applied when using non-linear function approximators as neural networks to represent the objectives in the game. Traditional algorithms average the policy, the mean-field distribution, or both, in order to guarantee a theoretical convergence to the MFNE. This can be done in the case of games with small state spaces under linear or tabular policies, but it is not straightforward when using neural networks. Recent works ~\cite{lauriere2022scalable} have derived deep learning algorithms based on value learning suitable for MFG. However, to the best of our knowledge, there is no approach based on policy optimization that addresses this issue.

The main contribution of this paper is bringing policy-based optimization into MFG. This is performed through developing an algorithm based on Proximal Policy Optimization (PPO)~\cite{schulman2017proximal}. We refer to this algorithm as \textit{Mean Field Proximal Policy Optimization (MF-PPO)}. Conducted experiments in the OpenSpiel framework~\cite{lanctot2019openspiel} show better convergence performance of MF-PPO compared with current state-of-the-art methods for MFG. This validates our approach and broadens the spectrum of algorithms on MFG to policy-based methods, traditionally dominant in the literature on environments with large or continuous action spaces.  

The remainder of this paper is organised as follows. 
In Section~\ref{sec::relatedWork}, we present the state-of-the-art related to solving the mean-field games. In Section~\ref{problem_formulation}, we provide a formal description of the problem formulation. Then, in Section \ref{sec::MF-PPO} we present the designed algorithm MF-PPO, that we validate experimentally in Section~\ref{sec::Exps}. Finally, Section~\ref{sec::conclusion} concludes the paper.

\footnote{Code available at: \url{https://github.com/Optimization-and-Machine-Learning-Lab/open_spiel/tree/master/open_spiel/python/mfg}}

\vspace{-1em}
\section{Related works}
\label{sec::relatedWork}
\vspace{-1em}
In the literature, numerous RL approaches have been designed to address MFG. These can be classified based on the property used to represent the population into (i) mean-field action and (ii) mean-field state distribution. Examples of mean-field action can be found in~\cite{subramanian2020partially}, in these works the interaction within the population is done based on the average behaviour of the neighbours. A more common approach is using the mean-field state distribution ~\cite{angiuli2022unified}. This approach approximates the infinitum of agents by the state distribution or \textit{distribution flow} of the population. In this case, each player is affected by other players through an aggregate population state.
Also, regarding the problem setup, MFG can be classified as (i) stationary or (ii) non-stationary. In the stationary setup, the mean field distribution does not evolve during the episode~\cite{subramanian2019reinforcement}.
A more realistic scenario, and the one discussed in this work, is the non-stationary~\cite{mishra2020model}. In that case the mean-field state is influenced by the agents decisions.

The methodology to address MFG in the literature is also diverse. The classical method for learning the MFNE is the (BP) algorithm~\cite{huang2006large}. BP is a fixed point iteration method that iteratively computes the Best Response (BR) for updating the mean field distribution. The convergence of the BP algorithm is restrictive~\cite{cui2021approximately}, and in practice, it might appear with oscillations. To address this issue, the Fictitious Play (FP) algorithm~\cite{cardaliaguet2017learning} averages the mean field distribution over the past iterations. This stabilizes the learning and improves the convergence properties of the algorithm~\cite{perrin2020fictitious}. 
Several attempts have been made in the literature to scale FP. For example, ~\cite{heinrich2016deep} proposed Neural Fictitious Self-play algorithm based on fitted Q-learning that learns from best response behaviours on previous experiences. Also, Deep Average Fictitious Play~\cite{lauriere2022scalable} presents a similar idea in a model-free version of FP in which the BR policy is learned though deep Q-learning. Although learning the best response using deep RL allows scaling this method to games with larger state spaces, in practice learning the BR policy is computationally inefficient. In this sense, algorithms based on policy iteration have been also applied to MFG~\cite{cacace2021policy}. These methods have proved to be more efficient~\cite{perolat2021scaling} as they do not require the computation of the best response but they perform a policy update per evaluation step. An example is Online Mirror Descent (OMD)~\cite{shalev2012online}, which averages the evaluation of the Q-function from where it derives the mean-field policy. A deep learning variant of it is the Deep-Munchausen OMD (D-MOMD) ~\cite{lauriere2022scalable}. This algorithm uses the Munchausen algorithm~\cite{vieillard2020munchausen} to approximate the cumulative Q-function when parameterized using a neural network.

\begin{table}[!t]
\centering
\caption{Summary on the RL literature for MFG.}
\begin{tabular}{l@{\hskip 0.2in}c@{\hskip 0.2in}c@{\hskip 0.1in}c@{\hskip 0.1in}c}
\hline \hline 
 & Setting & Learning & Requires Oracle & Best Response \\ \hline
Heinrich et al.~\cite{heinrich2016deep}  &   General RL     &  Value-based      &    Yes   & Yes  \\ 
Lauri{\`e}re et al.~\cite{lauriere2022scalable}  &   General RL     &  Value-based      &    Yes   & No  \\
Koppel et al.~\cite{koppel2022oracle}      &   General RL      &  Value-based      &    No    &  No \\
Xie et al.~\cite{xie2021learning}                    & General RL               & Value-based      &   No       & No   \\
Fu et al.~\cite{fu2019actor}                    & LQR               & Policy-based      &   No       & Yes   \\
\textbf{Our Approach}                                   & General RL        & Policy-based      &   Yes    &  No \\
\hline 
\end{tabular}
\label{tab:table1}
\end{table}

Last but not least, oracle-free methods~\cite{angiuli2021reinforcement} are complete model-free RL methods applied to MFG. Oracle-free algorithms do not require the model dynamics but they estimate the mean-field distribution induced by the population. In~\cite{koppel2022oracle}, the authors propose a two timescale approach with a Q-learning algorithm suitable for both cooperative and non-cooperative games that simultaneously updates the action-value function and the mean-field distribution in a two timescale setting.

Regardless of the numerous works on value-based learning, the attention to policy optimization methods in MFG has been limited. Related works cover the linear quadratic regulator setting~\cite{uz2020reinforcement} but not general RL, a summary can be observed in Table~\ref{tab:table1}. Motivated by~\cite{yu2021surprising}, work that emphasizes the effectiveness of PPO in multi-agent games, this paper brings PPO into MFG by presenting a solution to the stabilization issues based on proximal policy updates.


\vspace{-1em}

\section{Problem formulation}
\label{problem_formulation}
\vspace{-0.5em}
In Mean Field Games (MFG) the interaction between multiple agents is reduced to a uniform and homogeneous population represented by the mean-field distribution. This is the distribution over states that the continuum of agents define when following the mean-field policy. The way in which MFG addresses the problem is selecting a~\textit{representative player} that interacts with the mean-field distribution. This simplifies the problem and facilitates the computation of the equilibria.

More formally, we consider the non-stationary setting with a finite time horizon in which we denote by $n \in \{0,1,...,N_T\}$ the time steps in an episode. The state and actions of an agent at each time-step are denoted as $s_n \in \mathcal{S}$ and $a_n \in \mathcal{A}$, both finite in our setting. The mean-field state is represented by the distribution of the population states $\mu_n \in \Delta^{|{\mathcal{S}}|}$, where $\Delta^{|{\mathcal{S}}|}$ is the set of state probability distributions on $\mathcal{S}$. In the non-stationary setting, the mean field distribution $\mu_n$ evolves during the episode and it characterizes the model dynamics $P: \mathcal{S} \times \mathcal{A} \times \Delta^{|{\mathcal{S}}|} \rightarrow \Delta^{|{\mathcal{S}}|}$ and the reward function $R: \mathcal{S} \times \mathcal{A} \times \Delta^{|{\mathcal{S}}|} \rightarrow \mathbb{R}$. 
The policy of the agents depends on a prior on the mean-field distribution. Although, without loss of generality, we can define a time-dependent policy $\pi_n \in \Pi:\mathcal{S} \rightarrow \Delta^{|{\mathcal{A}}|}$ that independently reacts to the mean-field state at every step. The model dynamics are therefore expressed as
\vspace{-0.5em}
\begin{equation}
s_{n+1} \sim P(\cdot | s_n, a_n, \mu_n) \quad \quad a_{n} \sim \pi_n(\cdot|s_n).
\end{equation}

We define the policy $\boldsymbol{\pi}  :=(\pi_n)_{n\geq0}$ as the aggregated policy for every time-step, similarly the mean-field distribution $\boldsymbol{\mu}:=(\mu_n)_{n\geq0}$. The value function is calculated as $V^{\boldsymbol{\pi},\boldsymbol{\mu}}(s) := \mathbb{E}[\sum^{N_T}_{n=0}\gamma^n r(s_n,a_n,\mu_n)]$. Given a population distribution $\boldsymbol{\mu}$ the objective for the representative agent is to learn the policy $\boldsymbol{\pi}$ that maximizes the expected total reward,
\vspace{-0.5em}
\begin{equation}
\begin{gathered}
    J(\boldsymbol{\pi},\boldsymbol{\mu}) = \mathbb{E}_{a_n \sim \pi_n(\cdot|s_n) , s_{n+1} \sim P(\cdot | s_n, a_n, \mu_n)} \left[ \sum_{n=0}^{N_T} \gamma^n R(s_n,a_n, \mu_n) \: | \: \mu_{0} \sim m_{0} \:  \right]\\
\end{gathered}
\end{equation}
\noindent where $\mu_0$ is the initial mean-field state drawn from the initial distribution of the population $m_0$ and $0 < \gamma < 1$ denotes the discount factor.
 
\vspace{0.5em}
\noindent \textbf{Nash equilibrium in MFG.} The desired solution in games is computing the Nash Equilibrium. This is the set of policies that followed by all players maximize their individual reward such that no agent can unilaterally increase deviating from the Nash policy. Furthermore, in MFG the agents share the same interest and an extension of the Nash equilibrium is needed. 

\begin{definition}
A mean-field Nash equilibrium ($\text{MFNE}$) is defined as the pair $(\pi^*,\mu^*)$ that satisfies the rationality principle $V^{\pi^*,\mu^*}(s) \geq V^{\pi,\mu^*}(s) \: \forall s,\pi$; and the consistency principle, $\mu^*$ is the mean-field state distribution induced by all agents following optimal policy $\pi^*$.
\end{definition}




\noindent \textbf{Mean-field Dynamics.} This work relies on an \textit{oracle} to derive the mean-field state. Given the initial mean-field distribution $\mu_0 = m_0$, the oracle uses the transition function $P$ to compute the mean-field distribution induced by the policy $\pi_n$ at each time step $n \in \left\lbrace 0,1,...,N_T \right\rbrace $,
\vspace{-0.5em}
\begin{equation}
   \mu_{n+1}(s') = \sum_{s,a \in \mathcal{S}\times \mathcal{A}} \mu_n(s) \pi_n(a|s)  P(s'|s,a,\mu_n)  \quad \forall s' \in \mathcal{S}.
\end{equation}

In a similar way, the policy is evaluated analytically by computing the expected total costs of the policy $\boldsymbol{\pi}$ under the mean field $\boldsymbol{\mu}$ as follows:
\vspace{-0.5em}
\begin{equation}
    J(\boldsymbol{\pi},\boldsymbol{\mu}) = \sum_{n=0}^{N_t} \sum_{s,a \in \mathcal{S}\times \mathcal{A}} \mu_n(s)  \pi_n(a|s)  R(s,a,\mu_n).
\end{equation}

\noindent \textbf{Exploitability.} The metric of choice for estimating the MFNE convergence is the exploitability. This metric is well known in game-theory~\cite{bowling2015heads,lanctot2009monte} and it characterizes the maximum increase in the expected reward a representative player can obtain deviating from the policy the rest of the population adopted.
The exploitability is obtained as follows:
\begin{equation}
    \phi(\boldsymbol{\pi},\boldsymbol{\mu}) = \max_{\boldsymbol{\pi'}} J(\boldsymbol{\pi'},\boldsymbol{\mu}) - J(\boldsymbol{\pi},\boldsymbol{\mu}).
\end{equation}
An interpretation of the exploitability is to consider it as a measure of how close the learned policy is to the MFNE. Small values of exploitability indicate less incentive for any agent to change its policy.

\section{Proposal: Proximal policy updates for MFG}
\label{sec::MF-PPO}

Learning in MFG is commonly achieved in the literature via fixed-point iteration~\cite{huang2006large}, where the set $\{(\pi_k,\mu_k)\}_{k\geq0}$ is recursively updated. Particularly, at iteration $k$ the best response policy for the MDP induced by $\mu_k$ is computed and the mean-field is updated $\mu_{k+1}$ as a result of the many agent following $\pi^{BR}_k$. Under the assumptions discussed in~\cite{huang2006large}, contraction mapping holds and the algorithm is proof to converge to a unique fixed point $\{(\pi^*,\mu^*)\}$. This problem corresponds to finding the optimal policy for an MDP induced by $\mu$, $\text{MDP}_{\mu}:=(\mathcal{S,A},P(\mu),R(\mu),\gamma)$. This can be solved using modern RL techniques that allow in practice to scale the method to large games. 


However, solving the BR is demanding, and in practice, it leads to instabilities in learning. In this paper, we aim to provide a solution to these instabilities by regularizing the updates in the mean-field policy. To this end, we bring the proximal policy updates developed in PPO~\cite{schulman2017proximal} into MFG.

Let start defining how PPO can be used to estimate the best response $\hat{\pi}^{BR}_\mu$ to the $\text{MDP}_\mu$. Based on the trajectories collected during the iteration $k$, one can perform policy optimization on the following objective function
\vspace{-0.5em}
\begin{equation}
\mathcal{J}^{PPO}_{\mu}(\theta) = \hat{E}_n \left[\min ( r_n \hat{A}_n, \; \text{clip} (r_n \pm \epsilon) \hat{A}_n )\right] \quad \quad  r_n(\theta) = \frac{\pi(\cdot|s_n;\theta)}{\pi(\cdot|s_n;\theta_\text{old})}
\end{equation}
\noindent where $\pi(a_n|s_n;\theta)$ is a stochastic policy,  $\pi(a_n|s_n;\theta_{old})$ is the policy before the update and $\hat{A}_n$ is an estimator of the advantage function at timestep $n$. $\hat{E}$ is the empirical expectation based in Monte-carlo rollouts. 
The theory behind PPO suggests relaxing the update on the policy to prevent large destructive updates by using a clip function applied on the ratio between the old policy and the current one. PPO imposes this constraint, forcing the ratio on the policy update $r_n(\theta)$ to stay within a proximal interval. This is controlled with the clipping hyperparameter $\epsilon$.

\begin{algorithm}[!t]
\caption{MF-PPO algorithm}\label{alg:MF-PPO}
\begin{algorithmic}

\STATE Initial policy parameters $\theta$, initial value function parameters $\phi$, initialize mean-field policy parameters $\theta^0 \leftarrow \theta$, initial mean-field distribution $\mu_{0} = m_0$

\renewcommand{\COMMENT}[2][.25\linewidth]{\leavevmode\hfill\makebox[#1][l]{(#2)}}
  
\FOR{iteration $k = 1,2,...,K$}
    \STATE Compute the mean-field distribution $\mu^{k}$ induced by the policy $\pi_{\theta^{k-1}}$
    \FOR{epsiode $e = 1,2,..., E$}
        \STATE Sample a minibatch of transitions: $\mathcal{D}_e=\{s_n,a_n,r_n,s_{n+1}\}$ by running the policy $\pi_n (\theta)$ on the game governed by the mean-field $\mu^k_n$.
        \STATE Compute the advantage estimate: $\hat{A}_n = G_n - \hat{V}_\phi(s_{n})$
        \STATE Update the policy network: $\theta^\star \leftarrow \argmax_\theta \mathcal{L}^{\text{MF-PPO}}(\theta)$  \eqref{loss-func}
        
        \STATE Update the value network: $\phi \leftarrow \argmin_\phi \hat{E}_n [ \; || G_n - \hat{V}_\phi(s_n)||^2\;]$ 
    \ENDFOR
    
    \STATE Update the mean-field policy parameters: $\theta^{k} \leftarrow \theta$
\ENDFOR
\RETURN $\mu^K$, $\pi_{\theta^K}$
\end{algorithmic}
\end{algorithm}


In this work, we extend the regularization of the policy updates to successive iterations on the MFG. We call the algorithm \textit{Mean-Field Proximal Policy Optimization (MF-PPO)} and it combines a double proximal policy regularization for the intra- and inter-iteration policy updates. This prevents the mean-field policy from having a large update between iterations, obtaining a smoothing effect that has previously been reported beneficial in value-based algorithms for MFG~\cite{lauriere2022scalable}. We denote the probability ratios for the intra- and inter-iteration policy updates as
\vspace{-0.5em}
\begin{equation}
     r^{\text{e}}_n(\theta)=\frac{\pi_n(a_n|s_n;\theta)}{\pi_n(a_n|s_n; \theta^e_{\text{old}})}
     \quad \quad  r^{\text{k}}_n(\theta)=\frac{\pi_n(a_n|s_n;\theta)}{\pi_n(a_n|s_n;\theta^{k}_{\text{old}})}
\end{equation}
\noindent where the superscript $k\in [1,K]$ refers to the iteration and the superscript $e \in [1,E]$ to the episode.

In order to derive an appropriate objective function for MFG we extend the objective function of the classical PPO by adding an additional term that limits the policy updates w.r.t. the previous iteration. We can think in this term as a proximal update that limits the divergence between iterations preventing the policy from reaching the BR at iteration $k$. The MF-PPO objective is therefore expressed as
\begin{multline} \label{loss-func}
\mathcal{L}^{\text{MF-PPO}}(\theta) = \hat{E}[ \alpha \min(r^{\text{e}}_n \hat{A}_n, \text{clip}(r_n^{\text{e}} \pm \epsilon_e) \hat{A}_n) \\ + (1-\alpha) \min(r^{k}_n \hat{A}_n, \text{clip}(r_n^{\text{k}} \pm \epsilon_k) \hat{A}_n)]
\end{multline}
\noindent where $0 < \alpha < 1$ balances the proximity of the policy between the inter and intra-iteration updates.

\section{Experimentation}
\label{sec::Exps}
In this section, we describe the experiments conducted to validate the proposed MF-PPO algorithm. We analyze the hyper-parameter selection and finally, we present the numerical results obtained against the state-of-the-art algorithms namely Deep-Munchausen Online mirror decent (D-MOMD) and Deep Average-Network Fictitious Play (D-ANFP)~\cite{lauriere2022scalable}.

\vspace{-1em}
\subsection{Experimental setup}


We opted for the OpenSpiel suite~\cite{lanctot2019openspiel} to benchmark the proposed algorithm in selected crowd modeling with congestion scenarios. Particularly the scenarios used for evaluation are:
\vspace{-0.2em}
\begin{itemize}
    \item[] \textbf{Four-rooms.}
A simple setup on a four-room grid with $10\times10$ states and a time horizon of $40$ steps. The agents receive a reward for navigating close to the goal located in the bottom right room while there also exists an adversion to crowded areas.
    \item[] \textbf{Maze.}
The maze is a more complex scenario with $20\times20$ states and a time horizon of 100 steps. In this setting, the agent must correctly steer through a complex maze to reach the goal while, similar to the previous case, evading congested areas.
\end{itemize}

\noindent In both environments the state-space is a two-dimension grid, where the state is represented by the agent’s current position. Furthermore, the action space consists of five discrete actions: up, down, left, right, or nothing. Those actions are always valid if the agent is confined within the boundaries. Finally, the reward signal is defined as: 
\vspace{-0.5em}
\begin{equation}
    r(s,a,\mu) = r_{\text{pos}}(s) + r_{\text{move}}(a,\mu(s)) + r_{\text{pop}}(\mu(s))
\end{equation}
where the first term measures the distance to the target, the second penalizes movement, and the last term is a penalty which encourages the agents to avoid crowded areas, and is given by the inverse of the concentration of the distribution at a particular state. 
  
\vspace{-1em}
\subsection{Numerical results}

\begin{figure*}[!t]
    \centering
    \begin{subfigure}[b]{0.475\textwidth}
        \centering
        \includegraphics[width=\textwidth]{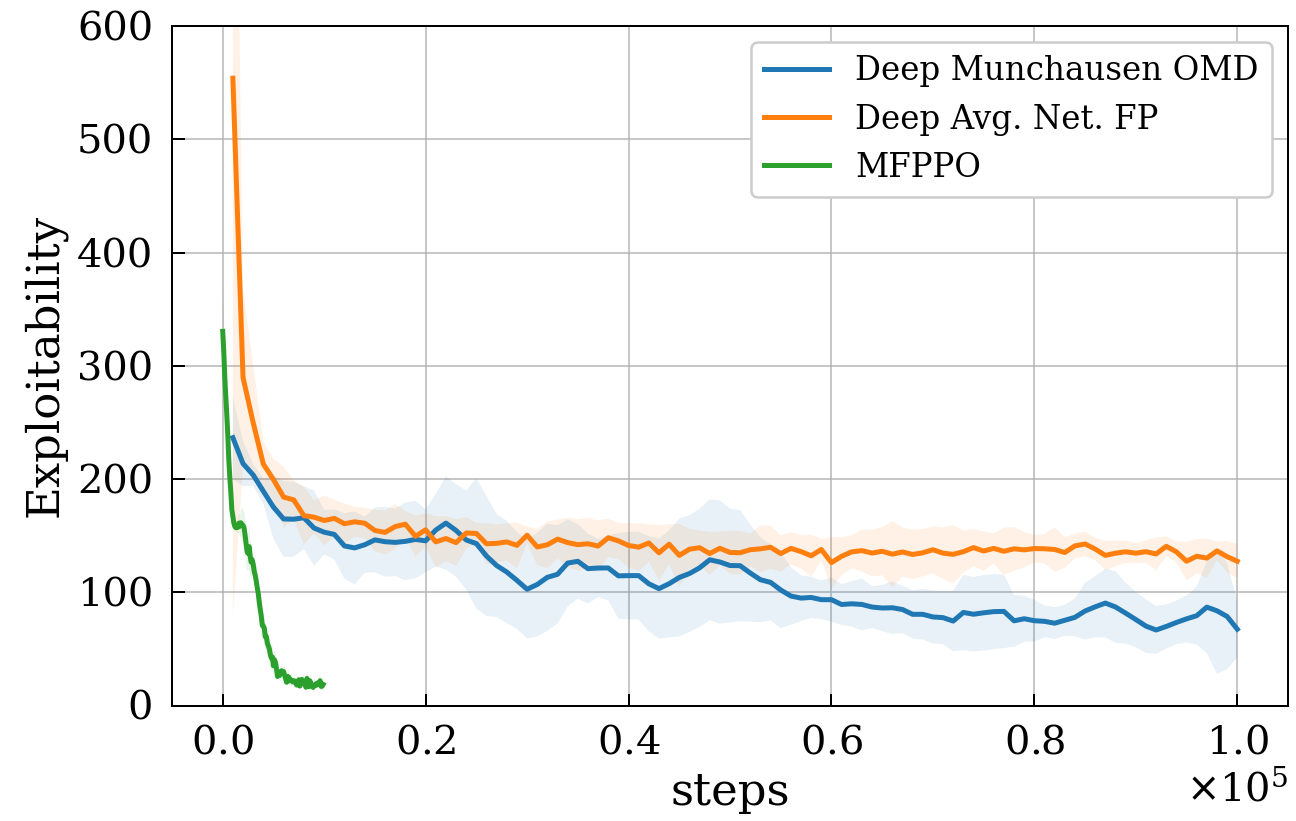}
        \caption{}
        \label{fig:mean and std of net14}
    \end{subfigure}
    \hfill
    \begin{subfigure}[b]{0.475\textwidth}  
        \centering 
        \includegraphics[trim={1.6cm 0 0 0},clip,width=\textwidth]{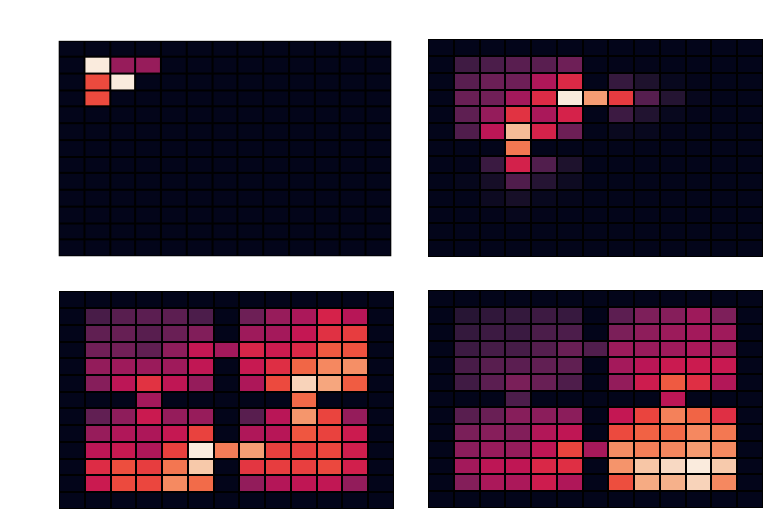}
        \caption{}  
        \label{fig:mean and std of net24}
    \end{subfigure}
    \vskip\baselineskip
    \begin{subfigure}[b]{0.475\textwidth}   
        \centering 
        \includegraphics[width=\textwidth]{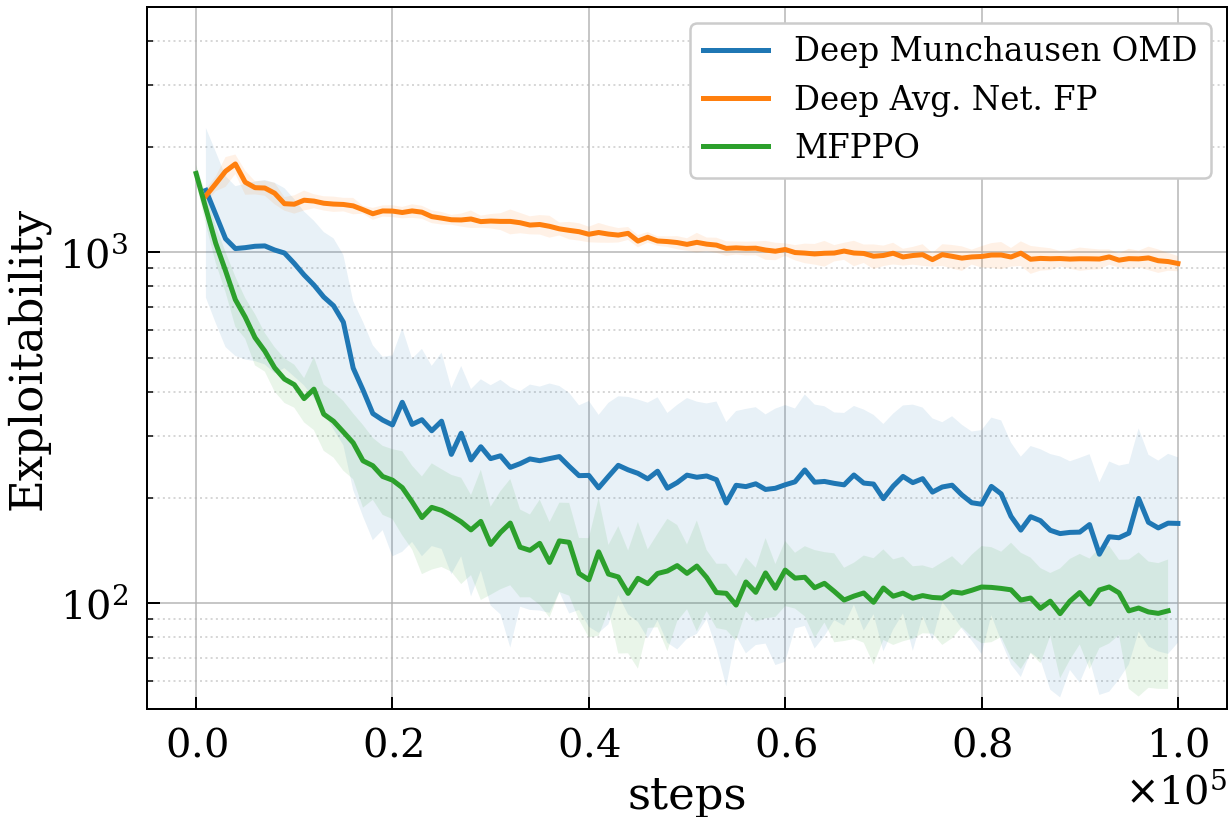}
        \caption{} 
        \label{fig:mean and std of net34}
    \end{subfigure}
    \hfill 
    \begin{subfigure}[b]{0.475\textwidth}   
        \centering 
        \includegraphics[width=\textwidth]{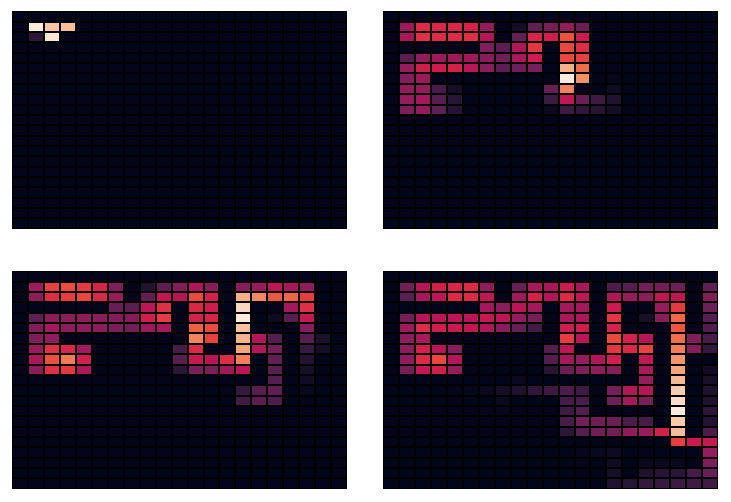}
        \caption{}
        \label{fig:mean and std of net44}
    \end{subfigure}
    \caption{\small On the left, the exploitability results obtained on the (a) four-room and (c) maze environments. Results are averaged over five seeds and the confidence interval corresponds to one standard deviation. On the right, the mean-field distribution of the agents generated by the MF-PPO policy on the (b) four-room and (d) maze environments.} 
    \label{fig:results}
\end{figure*}

In this section, we present the results MF-PPO achieves in the selected scenarios. We compare our results with Deep-Munchausen Online Mirror Descent (D-MOMD) and Deep Average-Network Fictitious Play (D-ANFP)~\cite{lauriere2022scalable}, both state-of-the-art algorithms in the selected settings. We report the exploitability metric, which is used in the literature as a proxy for quantifying convergence to the MFNE. The results are depicted in Fig.~\ref{fig:results} and summarized in Table~\ref{table:results}.


%

\subsubsection{Four-rooms.} Obtained results show that MF-PPO outperforms D-NAFP and D-MOMD algorithms, not only by converging to a better $\epsilon$-MFNE solution but, as depicted in Fig.~\ref{fig:results}, converging in a significantly fewer number of steps. We speculate that this can be credited to the fact our solution learns the optimal policy directly which in this situation is superior to learning the value function that the other methods use and then extract the optimal policy. Fig.~\ref{fig:results}(b) shows the learned mean-field distribution learned using MF-PPO. The agents gather as expected around the goal state at the right-bottom room, reaching it by equally distributing over the two symmetric paths.

\vspace{-1em}
\subsubsection{Maze.}
Similarly, on the Maze environment Fig.~\ref{fig:results}(c) shows that MF-PPO and D-MOMD converge to a favorable $\epsilon$-MFNE solution, whereas D-ANFP does to a sub-optimal solution. Still, the policy learned by MF-PPO is closer to the MFNE, reported by a smaller exploitability. Finally, Fig.~\ref{fig:results}(d) corroborates that the flow of agents over the maze distribute around the goal located in the lower right part of the maze.

\begin{table}[!t] 
\centering 
\captionof{table}{Comparison of the exploitability metric of the different algorithms. Results are averaged over five different seeds and reported as mean $\pm$ std.}
\begin{tabular}{l@{\hskip 0.2in}c@{\hskip 0.2in}c@{\hskip 0.1in}} 
\hline\hline  
Environment & Four Rooms & Maze\\ [0.5ex] 







\hline 
D-MOMD &	\multicolumn{1}{l}{64.41 $\pm$ 24.84}              & \multicolumn{1}{l}{153.80 $\pm$ 93.05}               \\
D-ANFP &	\multicolumn{1}{l}{127.37 $\pm$ 15.19 }              & \multicolumn{1}{l}{929.54  $\pm$ 46.36}               \\
\textbf{MF-PPO} &	\textbf{15.84}  $\pm$  1.95    & \textbf{93.63}  $\pm$  38.11    \\
\hline 
\end{tabular}
\label{table:results}
\end{table}
\begin{table}[!t] 
\centering 
\captionof{table}{Comparison of the CPU execution time of the different algorithms. Results are averaged over five different seeds and reported as mean $\pm$ std.}

\begin{tabular}{l@{\hskip 0.2in}c@{\hskip 0.2in}c@{\hskip 0.1in}} 
\hline\hline  
Environment & Four Rooms & Maze\\ [0.5ex] 
\hline 
D-MOMD &	\multicolumn{1}{l}{3H48M $\pm$ 1.79  Min}              & \multicolumn{1}{l}{7H35M $\pm$ 1.53 Min}               \\
D-ANFP &	\multicolumn{1}{l}{8H35M $\pm$ 56.36 Min }              & \multicolumn{1}{l}{9H45M $\pm$ 2.24 Min}               \\
\textbf{MF-PPO} &	\textbf{33M32S}  $\pm$  16.58 Sec   & \textbf{5H36M}  $\pm$  3.37 Min    \\
\hline 
\end{tabular}
\label{table:CPU-Time}
\end{table}

\vspace{1em}
In Table~\ref{table:CPU-Time} we present the CPU execution time of the tested algorithms. In all experiments we used AMD EPYC 7742 64-Core server processor to produce presented results. We note that the official implementation of D-ANFP and D-MOMD was used to reproduce previously presented results. MF-PPO coverages faster than both approaches, more notably, as evidenced by Fig.~\ref{fig:results}(a) in the four rooms case, MF-PPO converges within roughly 34 minutes compared to hours by the other two methods. We see a similar, although not as remarkable, trend in the maze as well, where MF-PPO converges in roughly five and half hours to a better MFNE point in comparison with the other techniques.

\subsection{Analysis on the hyper-parameters}

\begin{figure}[!t]
     \centering
     \begin{subfigure}[b]{0.32\textwidth}
         \centering
         \includegraphics[width=\textwidth]{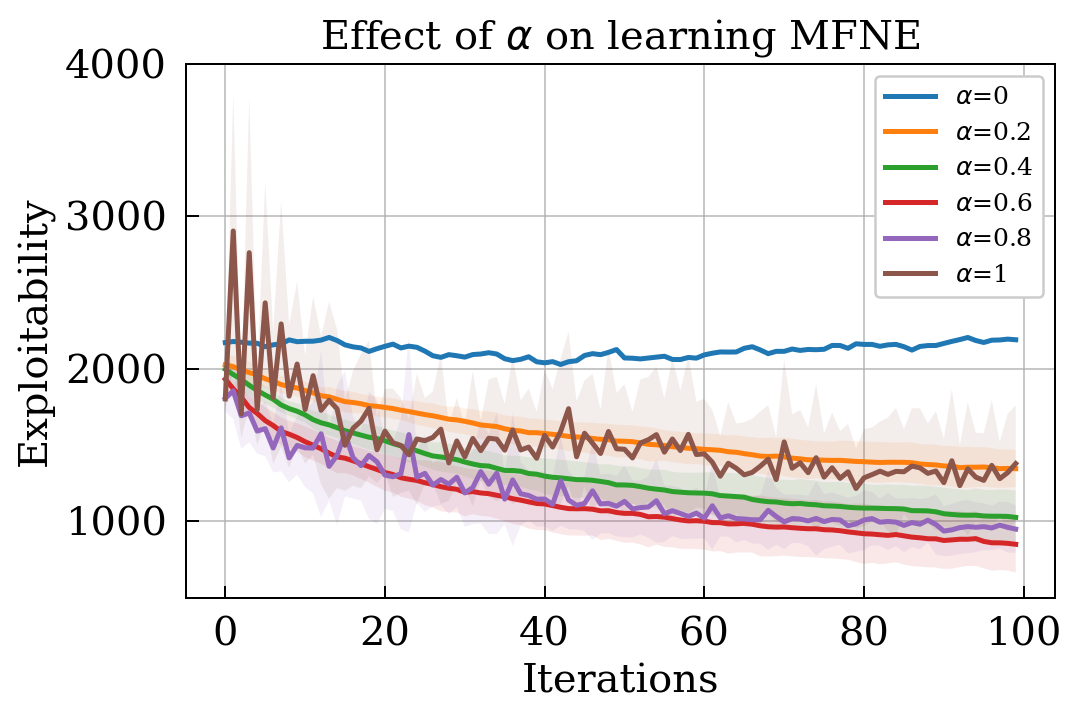}
        \caption{}
         \label{fig:alpha_beta}
     \end{subfigure}
     \hfill
     \begin{subfigure}[b]{0.32\textwidth}
         \centering
         \includegraphics[width=\textwidth]{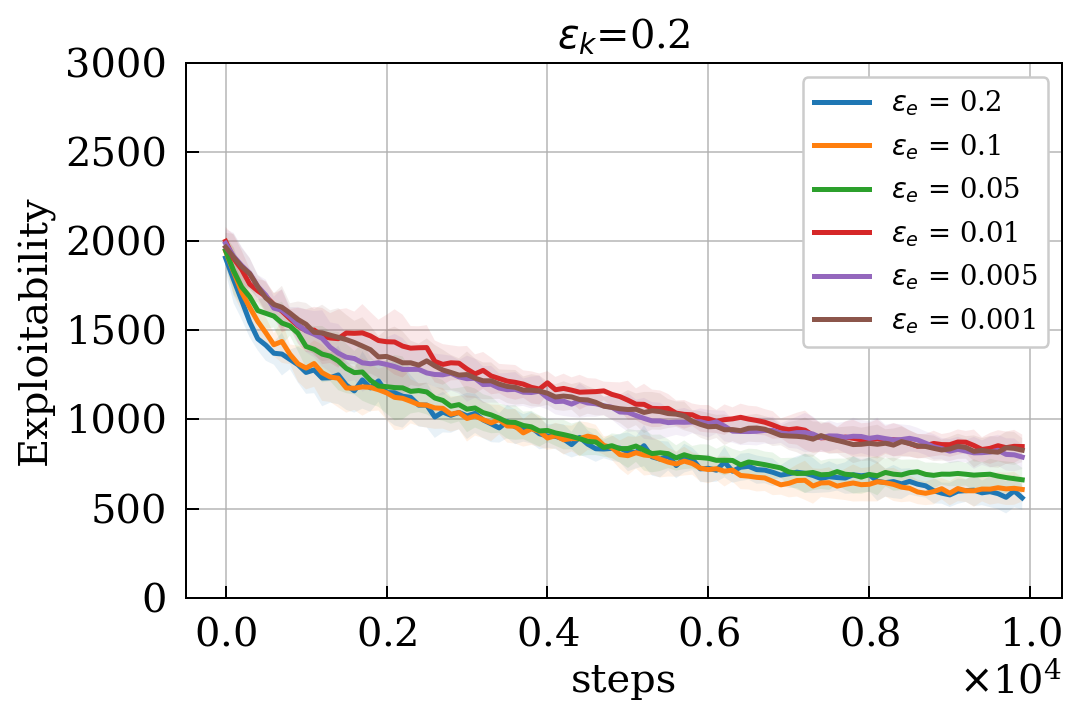}
        \caption{}
         \label{fig:First_eps}
     \end{subfigure}
     \hfill
     \begin{subfigure}[b]{0.32\textwidth}
         \centering
         \includegraphics[width=\textwidth]{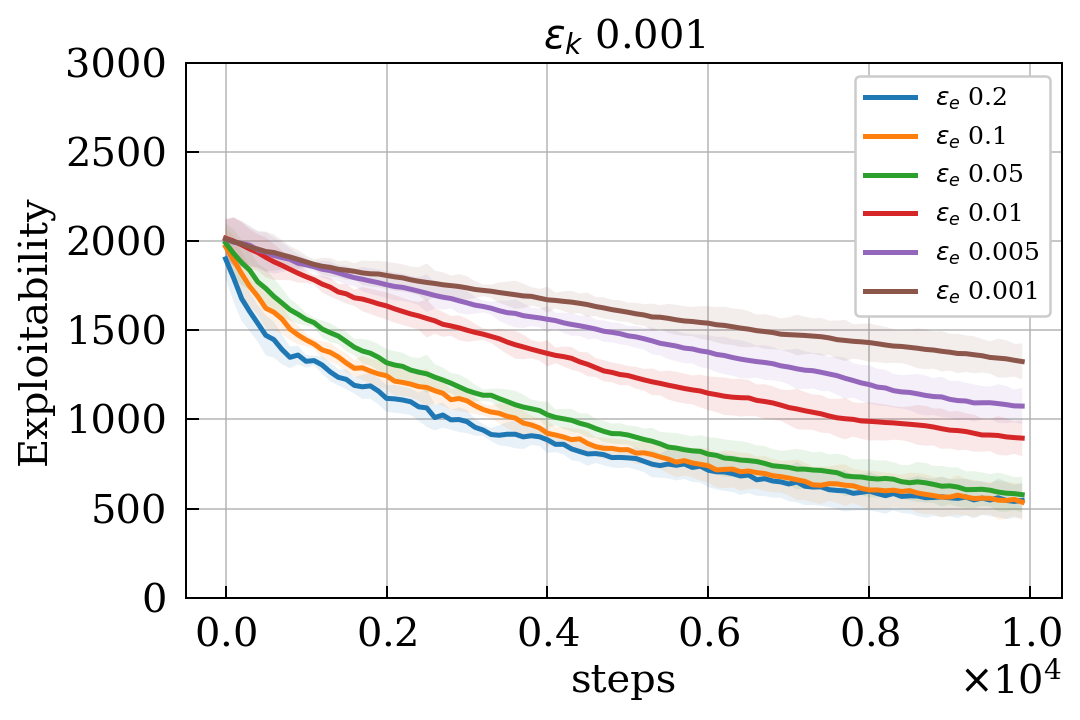}
        \caption{}
         \label{fig:sec_eps}
     \end{subfigure}
        \caption{ (a) Study on the impact of the hyper-parameter $\alpha$ on the learning. Moreover, (b) and (c) show the iteration clipping factor $\epsilon_k$ contribution to the smoothness and convergence of the MF-PPO algorithm.}
        \label{fig:Hyperparameter_study}
\end{figure}

This section investigates the influence on the hyper-parameter selection in the learning process. The experiments are conducted on the Maze environment. First, we focus on the configuration where $\alpha=0$, i.e., we update the iteration policy only and neglect the episode updates entirely. The results are depicted in Fig.~\ref{fig:Hyperparameter_study}(a), we see no sign of convergence indicated by high exploitability throughout learning. Furthermore, as the value of $\alpha$ assigned to episode updates increases, we observe a significantly better convergence rate. Nevertheless, it introduces oscillations that impede good convergence on the MFNE. This could be explained by the following dilemma: at each iteration, the representative agent learns an policy far better than what is available to the current population. Hence, agents have the incentive to deviate from the current policy resulting in an increment in the exploitability. Moreover, at the next iteration, the distribution is updated with such policy, resulting in a sharp decline in the exploitability. This phenomenon can be smoothed by reducing the rate at which the agent`s policy updates with respect to the mean-field policy. Consequently, this is controlled using the parameter $\alpha$ as shown by the remaining curves in Fig.~\ref{fig:Hyperparameter_study}(a).

\begin{table}[h] 
\centering 
\captionof{table}{Hyper-parameter selection.}
\vspace{8px}
\begin{tabular}{l@{\hskip 0.2in}c@{\hskip 0.2in}c@{\hskip 0.1in}}
\hline\hline  
Environment & Four Rooms & Maze\\ [0.5ex] 
\hline 
Input dimension	& 81	& 145\\
Critic/Actor network size	& [32, 32]	& [64, 64]\\
Critic output dimension	& \multicolumn{2}{c}{1}\\
Actor output dimension	& \multicolumn{2}{c}{5}\\
Activation function	& \multicolumn{2}{c}{ReLU}\\
Alpha $\alpha$	& 0.5	& 0.6\\
Iteration $\epsilon_{k}$	& 0.01	& 0.05\\
Episode $\epsilon_{e}$ & \multicolumn{2}{c}{0.2 }\\
Learning rate	& 1E-03& 	6E-04\\
Optimizer	& \multicolumn{2}{c}{Adam }\\
Update iteration 	& \multicolumn{2}{c}{100}\\
Update episodes	& 20	& 200\\
Update epochs	& \multicolumn{2}{c}{5}\\
Batch size 	& 200	& 500\\
Number of mini-batches	& 5& 	4\\
Gamma $\gamma$	& 0.99	& 0.9\\
\hline 
\end{tabular}
\label{table:hyperparameters}
\end{table}

Then we analyze the impact on both iteration $\epsilon_{k}$ and episode $\epsilon_{e}$ clipping factors. We consider two extreme cases for $\epsilon_{k}$ and different values of $\epsilon_{e}$. In Fig.~\ref{fig:Hyperparameter_study}(b), we set $\epsilon_{k} = 0.2$, and compare for different $\epsilon_{e}$ values, we observe high variance in exploitability mainly due to more significant policy updates. On the other end, for $\epsilon_{k}$ = 0.001, the curves look much smoother since the policy update is largely constrained; however, the drawback is a slower convergence rate as shown in Fig.~\ref{fig:Hyperparameter_study}(c).  Finally, all the hyper-parameters used in the experiments are summarized in Table \ref{table:hyperparameters}.


\section{Conclusion}
\label{sec::conclusion}
In this work, we propose the \textit{Mean Field Proximal Policy Optimization (MF-PPO)} algorithm for mean field games (MFG). Opposed to current strategies for stabilizing MFG based on averaging the q-values or the mean-field distribution, this work constitutes the first attempt for regularizing the mean-field policy updates directly. Particularly, MF-PPO algorithm regularizes the updates between successive iterations in the mean-field policy updates using a proximal policy optimization strategy. Conducted experiments in the OpenSpiel framework show a faster convergence to the MFNE when compared to current state-of-the-art methods for MFG, namely the Deep Munchausen Online Mirror Descent and Deep Average-Network Fictitious Play. 

As future work, the first track would be the investigation of the mathematical analysis of the MFNE reached by the MF-PPO algorithm. Second, investigating the optimization of the computation time of the proposed approach is of interest. Finally, the application of the approach on large-scale real cases would push the boundaries of the approach. 

\bibliographystyle{unsrt}
\bibliography{references}

\end{document}